\documentclass[10pt,twocolumn,letterpaper]{article}

\usepackage{cvpr}              

\usepackage{graphicx}
\usepackage{amsmath}
\usepackage{amssymb}
\usepackage{booktabs}
\usepackage{multirow}
\usepackage[dvipsnames]{xcolor}
\usepackage[accsupp]{axessibility} 

\usepackage{glossaries}
\newacronym{PnP}{PnP}{Perspective-n-Point}
\newacronym{APR}{APR}{Absolute Pose Regression}
\newacronym{OOD}{OOD}{Out-of-Domain}

\usepackage{cellspace}
\usepackage{tikz}

\definecolor{cvprblue}{rgb}{0.21,0.49,0.74}
\usepackage[breaklinks,colorlinks,allcolors=cvprblue]{hyperref}

\usepackage[capitalize]{cleveref}
\crefname{section}{Sec.}{Secs.}
\Crefname{section}{Section}{Sections}
\Crefname{table}{Table}{Tables}
\crefname{table}{Tab.}{Tabs.}

\global\long\def\comma{\enspace\mbox{,}}%

\def\cvprPaperID{53} 
\def\confName{CVsports}
\def\confYear{2025}
\newcommand{\mysection}[1]{\vspace{2pt}\noindent\textbf{#1}}

\usepackage{xcolor, color, colortbl}         
\definecolor{Highlight}{HTML}{39b54a}  

\usepackage{amssymb}
\usepackage{pifont}

\usepackage{algorithm}
\usepackage{listings}
\usepackage{etoolbox}
\makeatletter
\AfterEndEnvironment{algorithm}{\let\@algcomment\relax}
\AtEndEnvironment{algorithm}{\kern2pt\hrule\relax\vskip3pt\@algcomment}
\let\@algcomment\relax
\newcommand\algcomment[1]{\def\@algcomment{\footnotesize#1}}
\renewcommand\fs@ruled{\def\@fs@cfont{\bfseries}\let\@fs@capt\floatc@ruled
  \def\@fs@pre{\hrule height.8pt depth0pt \kern2pt}%
  \def\@fs@post{}%
  \def\@fs@mid{\kern2pt\hrule\kern2pt}%
  \let\@fs@iftopcapt\iftrue}
\makeatother
\newcommand{\cmmnt}[1]{}

\usepackage[normalem]{ulem}

\usepackage{listings}
\usepackage{color}
\definecolor{codegreen}{rgb}{0,0.6,0}
\definecolor{codegray}{rgb}{0.5,0.5,0.5}
\definecolor{codepurple}{rgb}{0.58,0,0.82}
\definecolor{backcolour}{rgb}{1,1,1}
 
\lstdefinestyle{mystyle}{
    backgroundcolor=\color{backcolour},   
    commentstyle=\color{codegreen},
    keywordstyle=\color{magenta},
    numberstyle=\tiny\color{codegray},
    stringstyle=\color{codepurple},
    basicstyle=\footnotesize,
    breakatwhitespace=false,         
    breaklines=true,                 
    captionpos=b,                    
    keepspaces=true,                 
    numbers=left,                    
    numbersep=5pt,                  
    showspaces=false,                
    showstringspaces=false,
    showtabs=false,                  
    tabsize=2
}
 
\makeatletter

\newcommand{\Rmnum}[1]{\expandafter\@slowromancap\romannumeral #1@}
\makeatother

\usepackage{afterpage}
\newcommand{\flushfloats}{
  \ifvmode
  \ifnum \@dbltopnum = \m@ne
    \ifdim \pagetotal <\topskip \hbox{} \fi
    \fi
    \fi
}

\begin{document}

\title{Can Geometry Save Central Views for Sports Field Registration?}

\author{Floriane Magera$^{1,2}$%
\quad
Thomas Hoyoux$^{1}$
\quad
Martin Castin$^{1}$  
\quad
Olivier Barnich$^{1}$  
\\
\quad
Anthony Cioppa$^{2}$
\quad
Marc Van Droogenbroeck$^{2}$  
\and $^1$ {\small EVS Broadcast Equipment}
\quad $^2$ {\small University of Li{\`e}ge, Belgium}
}

\maketitle

\global\long\def\RR{\mathbb{R}}%
\global\long\def\PP{\mathbb{P}}%

\global\long\def\firstCoordinateEuclidenSpace{x}%
\global\long\def\secondCoordinateEuclidenSpace{y}%
\global\long\def\vector#1{\textbf{\text{#1}}}%
\global\long\def\lineInProjectiveSpace#1{\vector{#1}}%
\global\long\def\genericLine{\lineInProjectiveSpace l}%
\global\long\def\linf{\lineInProjectiveSpace{\genericLine}_{\text{inf}}}%
\global\long\def\lvanishing{\lineInProjectiveSpace{\genericLine}_{\text{vanishing}}}%
\global\long\def\transpose#1{#1^{\mathsf{T}}}%
\global\long\def\inverse#1{{#1}^{-1}}%
\global\long\def\matrix#1{\mathtt{#1}}%
\global\long\def\identity{\matrix{\mathbf{I}}}%
\global\long\def\jaccard{\text{JaC}}%
\global\long\def\calibrationMetric{\jaccard}%
\global\long\def\calibrationMetricWithThreshold#1{\calibrationMetric_{#1}}%
\global\long\def\meanReprojectionError{\text{MRE}}%
\global\long\def\comma{\,\mbox{,}}%
\global\long\def\dot{\,\mbox{.}}%

\begin{abstract}
Single-frame sports field registration often serves as the foundation for extracting 3D information from broadcast videos, enabling applications related to sports analytics, refereeing, or fan engagement.  As sports fields have rigorous specifications in terms of shape and dimensions of their line, circle and point components, sports field markings are commonly used as calibration targets for this task. 
However, because of the sparse and uneven distribution of field markings, close-up camera views around central areas of the field often depict only line and circle markings. On these views, sports field registration is challenging for the vast majority of existing methods, as they focus on leveraging line field markings and their intersections. It is indeed a challenge to include circle correspondences in a set of linear equations. 
In this work, we propose a novel method to derive a set of points and lines from circle correspondences, enabling the exploitation of circle correspondences for both sports field registration and image annotation.  
In our experiments, we illustrate the benefits of our bottom-up geometric method against top-performing detectors and show that our method successfully complements them, enabling sports field registration in difficult scenarios. 
\end{abstract}

\section{Introduction}
\label{sec:intro}

In the realm of sports analytics, sports field registration methods are key to interpreting the game from the broadcast feeds. Accurately projecting any 3D real-world sport field point onto the 2D image plane enables numerous sport analytics applications. Indeed, extracting statistics about player performances~\cite{Tian2019UseOf}, actions such as goals, offsides, or fouls~\cite{Held2023VARS}, or the game state~\cite{Somers2024SoccerNetGameState} relies on the ability to locate the ball and players on the sports field~\cite{Fujii2025Machine}.

\begin{figure}[t]
\begin{center}
\includegraphics[width=1\linewidth]{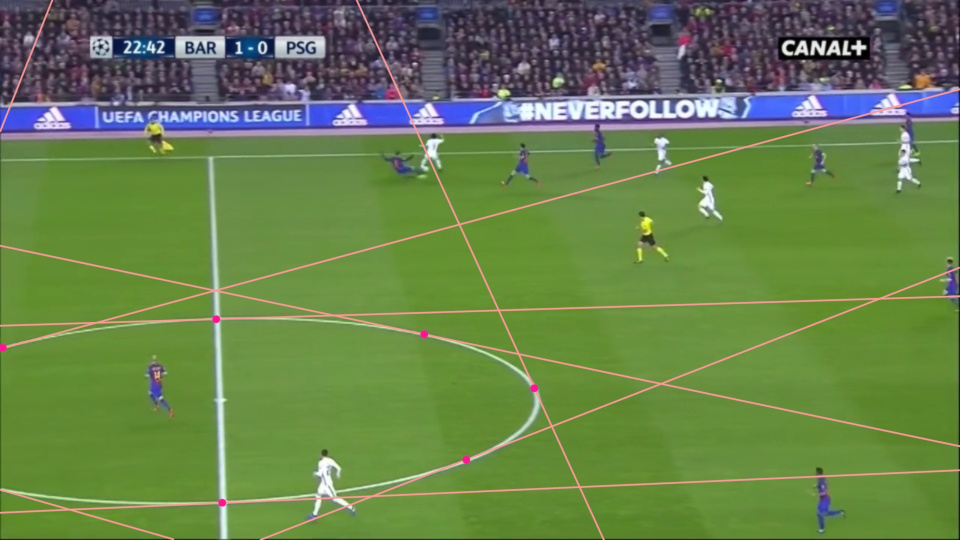}
\end{center}
\caption{Single-frame camera calibration is challenging when only a limited amount of field markings are visible in the scene. For example, in this image, camera parameters cannot be derived from the line and point correspondences  without incorporating priors about the camera. This is why current methods generate virtual keypoints (PnLCalib~\cite{Gutierrez2024Pnlcalib-arxiv} in yellow) which here, however, are not sufficient to allow for a registration of the sports field. Our geometric approach derives the pink points from the fitted ellipse and the detected center, which allow for the estimation of the homography (blue).
}\label{fig:GraphicalAbstract}
\end{figure}

An advantageous feature of sports fields is their fixed dimensions~\cite{FIFA2019Handbook,IIHF2022Official,FIBA2022Official}, which makes them convenient calibration targets. However, the sports field markings might be sparsely distributed in some areas, making the single-frame camera calibration particularly challenging, as shown in~\cref{fig:GraphicalAbstract}. For this reason, sports field registration, \ie homography estimation methods, are often considered as a proxy to camera calibration in sports. In fact, only four non-collinear point correspondences are needed to estimate the homography matrix that maps the sports field top-view template to its image, while four to six points are required to estimate the projection matrix of a pinhole camera depending on the \gls{PnP} solver used~\cite{Kukelova2016Efficient}. 

Nevertheless, four non-collinear point correspondences may not always be found directly in the image, as broadcast cameras often happen to zoom in on an action taking place nearby a mostly isolated circular marking on the sports field, usually around the center circle (as in \cref{fig:GraphicalAbstract}). Fortunately, we show that there are several options to handle those difficult views. 
In this work, we propose a solution that enables the registration of such zoomed-in camera views. For that, we take a geometrical, constructive approach that leverages the properties of circle projections to enable the retrieval of point and line correspondences.  

We will also explore a more difficult case, when only a small part of the circle is visible, by exploiting the constant thickness of the sports field markings on the ground.

\mysection{Contributions.} We summarize our contributions as follows:
\textbf{(i)} We introduce a novel method (relying on three variants depending on image support) to derive the set of primitives --points and lines-- that are necessary to estimate camera parameters for the challenging central view case. In this way, we are broadening the spectrum of scenes that can be calibrated.
\textbf{(ii)} Besides enabling the calibration of difficult views, our method offers a precise way to annotate point correspondences along circles.
Finally, \textbf{(iii)} we investigate the potential of our method on the SoccerNet dataset~\cite{Cioppa2022Scaling} and on challenging cases through qualitative and quantitative experiments.

\section{Related work}
\label{sec:related_work}

\begin{figure}
\begin{centering}
\includegraphics[width=1\columnwidth]{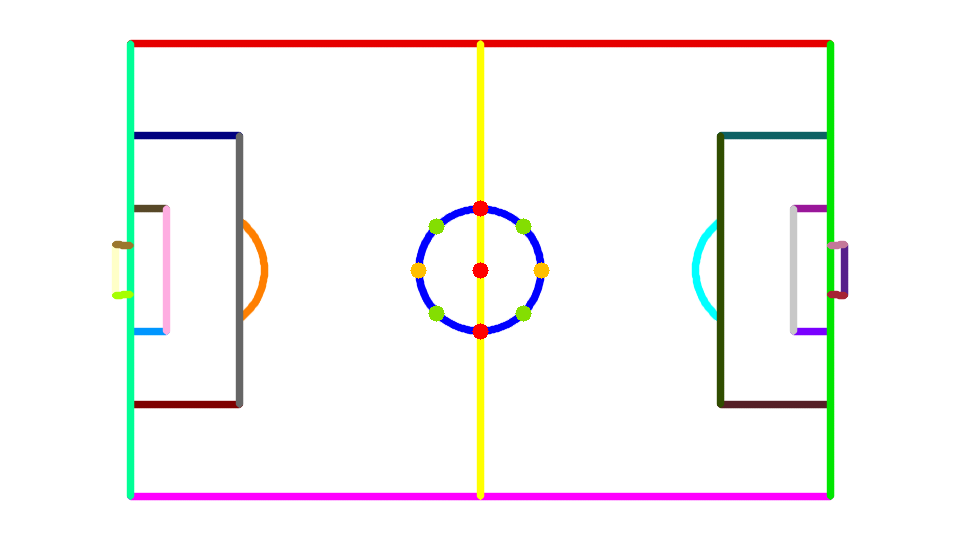}
\par\end{centering}
\caption{Top-view of the soccer field layout, as decomposed into different semantic elements. These elements form a set of markings include points, lines, and circles that can be used to calibrate a camera.   \label{fig:top-view-model}}
\end{figure}

With the recent SoccerNet Game State Recognition challenge~\cite{Somers2024SoccerNetGameState} and the Roboflow tutorial on soccer analytics~\cite{Skalski2024Camera}, a range of applications creating a radar view of the game with player tracking has flourished, with varying degrees of accuracy. A common flaw of these solutions is the visible shift of the players' position when the camera targets the central view of the soccer field, indicating a problem in soccer field registration for central views. 
All single-frame methods for camera calibration in sports heavily rely on the visibility of sports field markings in the camera view. Most methods leverage either the detection or the segmentation of such markings to derive the homography that maps the camera view to the template of the sports field (depicted in \cref{fig:top-view-model}). Difficulties arise from the fact that sports field markings are typically composed of a sparse collection of heterogeneous primitives, such as points, lines, and circles, unevenly distributed across the playfield. This issue makes camera calibration in sports challenging for narrow-angle camera views, especially for central views where only the halfway line and center circle are visible. 

Over the years, several methods have tried to solve the problems of sports field registration or camera calibration in sports by addressing the potential lack of visible markings in the image, as described hereafter.

\mysection{Dictionary-based methods.} A common approach is to enumerate all plausible camera views of the sports field markings in a dictionary. At detection time, the closest sample from the dictionary is retrieved to initialize the camera parameters and these parameters are later refined based on the field markings detection in the image~\cite{DAmicantonio2024Automated, Zhang2021AHigh, Sha2020EndtoEnd, Chen2019Sports, Sharma2018Automated}. These solutions usually assume that the set of camera parameters or homographies encountered in practice are well-known. They also tend to be quite slow in practice, as the size of the dictionary is proportional to the range of cameras to be calibrated \eg, 450 samples~\cite{Sha2020EndtoEnd}.

\mysection{Regression methods.} Starting from a first homography estimated with a deep neural network~\cite{Jiang2020Optimizing, Shi2022Self, Fani2021Localization-arxiv}, \ie learned homography priors, or from explicit camera parameters priors~\cite{Theiner2023TVCalib}, these methods refine their initial estimate using deep neural networks. While the initial work on which some of these methods build estimates the relative homography between frames~\cite{DeTone2016DeepImage-arxiv}, a direct sports field registration through homography regression is closer to an \gls{APR} task.

In their work, Sattler~\etal~\cite{Sattler2019Understanding} show that convolutional neural networks trained to solve the \gls{APR} task fail to generalize beyond their training sets, and  rather perform pose approximation via image retrieval than actual pose estimation. Given that sports datasets are scarce, consisting of few images, and generally only depict top-tier games, we fear these methods will not generalize well and will only work for top-tier main cameras. 
 
\mysection{Virtual augmentation of keypoints.} Some solutions augment identifiable sports field landmarks with additional virtual keypoints which are then used to train neural networks. These extra keypoints may be:
(1) a grid of regularly spaced keypoints of the whole sports field plane~\cite{Nie2021Robust, Chu2022Sports, Maglo2023Individual, Claasen2023Video}, or
(2) points on sports field circles, typically regularly spaced \eg, every 45 degrees~\cite{Falaleev2023Sportlight,Falaleev2024Enhancing, GutierrezPerez2024NoBells, Gutierrez2024Pnlcalib-arxiv}. 

Our work specifically addresses the potential shortcomings of these two approaches. First, those rely on the generalization capabilities of deep neural networks to detect points that do not have anything distinguishable but their location; there is no pattern to be learned for the keypoints mentioned in (1), they correspond to texture-less grass, and for the keypoints mentioned in (2), they correspond to a specific location along a curve, whose appearance is a function of the perspectivity. The ability of a neural network to detect those relies on its ability to grasp the perspectivity of the image, which rather relates to an \gls{APR} task than a detection or matching task. We hypothesize that the generalization capabilities of the neural networks detecting virtual keypoints may suffer from the same problem as the above-mentioned regression methods and thus be restricted to top-tier main cameras.
Second, these approaches need to register the sports field on these images to create annotations, which is always done through homography estimation. And, as previously mentioned, only images already containing a sufficient number of primitives can be annotated, while these additional points are needed the most when we have fewer than four point or line correspondences. There is thus a chicken-and-egg problem here. 

\mysection{Specific solutions for central views.}
Some previous work directly address circle correspondences in their method: 
Gupta~\etal~\cite{Gupta2011Using} leverage ellipses through their pole-polar relationship with already found point or line matches, thus enforcing geometric constraints to estimate homographies for hockey rink registration. Alvarez and Caselles~\cite{Alvarez2014Homography} derive a closed-form solution for homography estimation between a soccer field and an image using correspondences of the center circle, the halfway line, and up to three points belonging to this line.
Our method is similar in the sense of geometric exploitation of circle correspondences but targets a wider scope than homography estimation: the correspondences that we output also allow camera parameters estimation and keypoint annotation. 

\begin{figure}
\begin{centering}
\includegraphics[width=1\columnwidth]{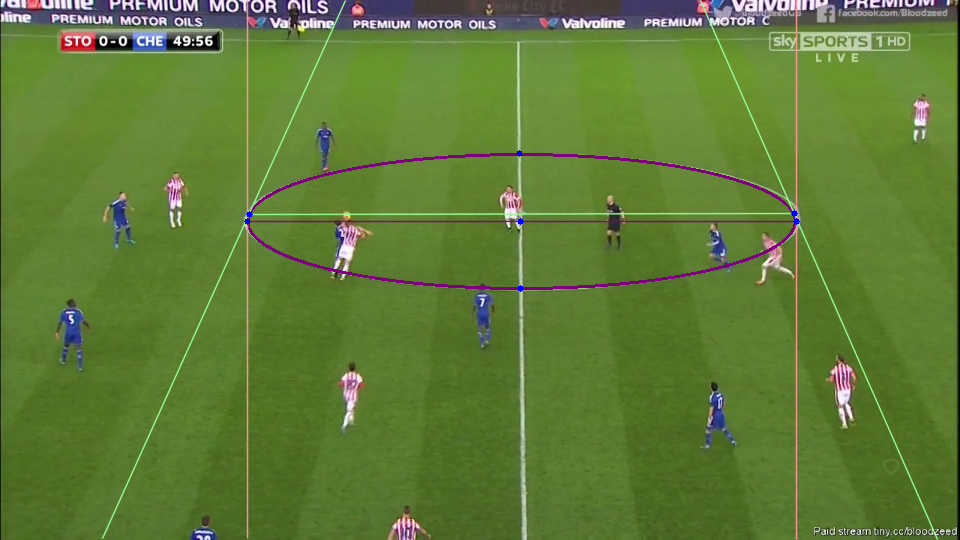}
\par\end{centering}
\caption{Visualization of the fitted ellipse (purple), its tangents from the great axis extremities (orange), or from the points extracted by our proposed method (green), that come from the line crossing the actual image of the circle center. The green tangents better capture the perspectivity as parallels should intersect at one point on the vanishing line of the supporting plane.  \label{fig:bad-tangents}}
\end{figure}

\mysection{Approximations with the great axis of the ellipse.} Other similar works make relatively strong assumptions about the type of perspectivity at hand.
Andrews~\etal~\cite{Andrews2024FootyVision} derive the vanishing point of the middle line and its parallels, then link it to the extremities of the ellipse great axis to derive additional point and line correspondences. Cuevas~\etal~\cite{Cuevas2020Automatic} match the great ellipse axis extremities with the circle points in orange as displayed in \cref{fig:top-view-model}, which is also the approach chosen in the Roboflow tutorial~\cite{Skalski2024Camera}. However, the projection of the circle center does not necessarily land on the ellipse center. Hence, the matching between the extremities of the great ellipse axis and the orange points in \cref{fig:top-view-model} is an approximation whose negative effect is depicted in \cref{fig:bad-tangents}. 

\mysection{Generic ellipse-based geometric methods.}
Beyond sports applications, generic solutions for homography estimation or camera calibration with ellipse correspondences have been studied. Chum and Matas~\cite{Chum2012Homography} linearize ellipse correspondences using a first-order Taylor expansion of the homography mapping around two or more ellipse correspondences.
Chen and Wu~\cite{Chen2004Camera} derive the focal length and extrinsic camera parameters from two coplanar circle correspondences.
Huang~\etal~\cite{Huang2016Homography} leverage the properties of the common self-polar triangle of separate ellipses to derive homographies. Finally, Huang~\etal~\cite{Huang2015TheCommon} recover the camera intrinsic parameters by leveraging concentric circles correspondences in three views. 

Inspired by generic solutions using ellipse correspondences, we investigated the use of a fully geometric bottom-up approach to derive point and line correspondences from circle ones, and their application to camera calibration to address the challenging central views in soccer.

\section{Method}
\label{sec:method}

\mysection{Preliminaries.} The goal of our method is to convert a circle correspondence into several point and line correspondences that can be integrated in the direct linear transform (DLT) usually done to estimate a homography or in a \gls{PnP} solver to extract camera parameters. Since many solutions to sports field markings segmentation and detection exist in the literature, we take off-the-shelf detectors, which are already capable of segmenting circular markings and detecting remarkable keypoints on sports fields, including the center of these circles if they are an integral part of the markings.

The sole detection of a circle does not suffice for calibration. Hence, starting from the detection of a circle correspondence, we developed three cases. The first two combine the circle correspondence and the detected projected circle center to: \textbf{Case 1} (\cref{section:line-correspondence}) a line correspondence, \textbf{Case 2} (\cref{section:pt-correspondence}) a point correspondence, and the last one, \textbf{Case 3} (\cref{sec:hidden-center}) is a fallback in case of unknown projection of the circle center. In the following, we start by defining the geometrical notions that will be necessary to derive a sufficient amount of linear constraints to estimate the homography. 

\paragraph{Geometric principles in the top-view.} Let us consider the supporting plane of the sports field $\pi$ from the 2D projective plane $\PP^2$ (notations and equations are taken from~\cite{Hartley2004Multiple}).  In the reference system of $\pi$, world geometric elements are described by their 2D homogeneous coordinates. Let $\matrix C$ be the world circle on the sports field, which is centered at location
$\vector o=\transpose{(o_{\firstCoordinateEuclidenSpace},o_{\secondCoordinateEuclidenSpace}, 1)}$, and has a radius $r$.

A point $\vector{x}$ expressed in homogeneous coordinates $\transpose{(\firstCoordinateEuclidenSpace, \secondCoordinateEuclidenSpace, 1)}$ belongs to the circle if
\begin{equation}
\transpose{\vector x}\matrix C\vector x = 0\comma
\end{equation}
where the matrix representation of $\matrix C$ is given by
 
\begin{equation}
\matrix C=\begin{bmatrix}1 & 0 & -o_{\firstCoordinateEuclidenSpace}\\
0 & 1 & -o_{\secondCoordinateEuclidenSpace}\\
-o_{\firstCoordinateEuclidenSpace} & -o_{\secondCoordinateEuclidenSpace} & o_{\firstCoordinateEuclidenSpace}^{2}+o_{\secondCoordinateEuclidenSpace}^{2}-r^{2}
\end{bmatrix}\dot
\end{equation}

A useful property of conics is the pole-polar relationship defined by a conic. It allows to map a point $\vector p$, the pole, and the \emph{polar} (line) $\genericLine$ with respect to conic $\matrix C$  by $\genericLine = \matrix C \, \vector p $ \cite[page 58]{Hartley2004Multiple}. 
The polar of the circle's center is the line at infinity of
the supporting plane $\pi$. Indeed,  
\begin{equation}
\begin{split}
\matrix C \, \vector o=\begin{bmatrix}1 & 0 & -o_{\firstCoordinateEuclidenSpace}\\
0 & 1 & -o_{\secondCoordinateEuclidenSpace}\\
-o_{\firstCoordinateEuclidenSpace} & -o_{\secondCoordinateEuclidenSpace} & o_{\firstCoordinateEuclidenSpace}^{2}+o_{y}^{2}-r^{2}
\end{bmatrix}\begin{pmatrix}o_{\firstCoordinateEuclidenSpace}\\
o_{\secondCoordinateEuclidenSpace}\\
1
\end{pmatrix}
\\ 
=\begin{pmatrix}0\\
0\\
-r^{2}
\end{pmatrix}=-r^{2}\begin{pmatrix}0\\
0\\
1
\end{pmatrix}=-r^{2}\,\linf\dot
\end{split}
\end{equation}
An important property of pole-polar relationships is their preservation through projective transformations. This means that the estimation of the imaged center of the circle in our image is key to deriving the projection of the vanishing line of $\pi$ in the image. 

Therefore, starting from the ellipse $\matrix E = \matrix H^{-T}\matrix C \,\matrix H^{-1}$ that corresponds to the projection of the circle, and the projection of its center: $\vector c = \matrix H \, \vector o$, we estimate the vanishing line:
\begin{equation}
\lvanishing = \matrix E \, \vector c\dot
\end{equation}
In the next subsection, we explain how to derive a sufficient number of correspondences based on one more primitive to estimate a homography matrix. 

\subsection{Case 1: Line correspondence}\label{section:line-correspondence}
Given a known line $\genericLine_1$ detected in the image, we can estimate the
vanishing point $\vector v$ linked to all parallels to $\genericLine _1$ by
\begin{equation}
\vector{v} =\genericLine_1\times \lvanishing \dot
\end{equation}
Once that $\vector v$ is known, it is easy to find the parallel to
$\genericLine_1$ passing through the projected circle center $\vector c$ : 
\begin{equation}
\genericLine_2=\vector{\vector v}\times \vector c\dot
\end{equation}
As illustrated in the left part of \cref{fig:derived_corresp}, this line $\genericLine_2$
intersects the ellipse in two points $\vector a$ and $\vector b$, which are our first two point correspondences. Then, we define the polar of $\vector v$, 
\begin{equation}
\genericLine_3 =\matrix E \, \vector v \comma
\end{equation}
which passes through the projected circle center $\vector c$, and intersects the ellipse in two new points $\vector d$ and $\vector e$ because $\vector v$ and $\vector c$ are conjugates. Furthermore, by definition, the polars
of $\vector d$ and $\vector e$ are tangents to the ellipse and intersect at $\vector v$; they are thus parallel to $\genericLine_1$, and perpendicular
to $\genericLine_3$. As the equation of the line $\genericLine_1$ is known in the sports field template, it is
direct to derive the 2D position of each derived point.

\subsection{Case 2: Point correspondence}\label{section:pt-correspondence}
\begin{figure*}[tb]
\begin{center}
\includegraphics[width=0.92\linewidth]{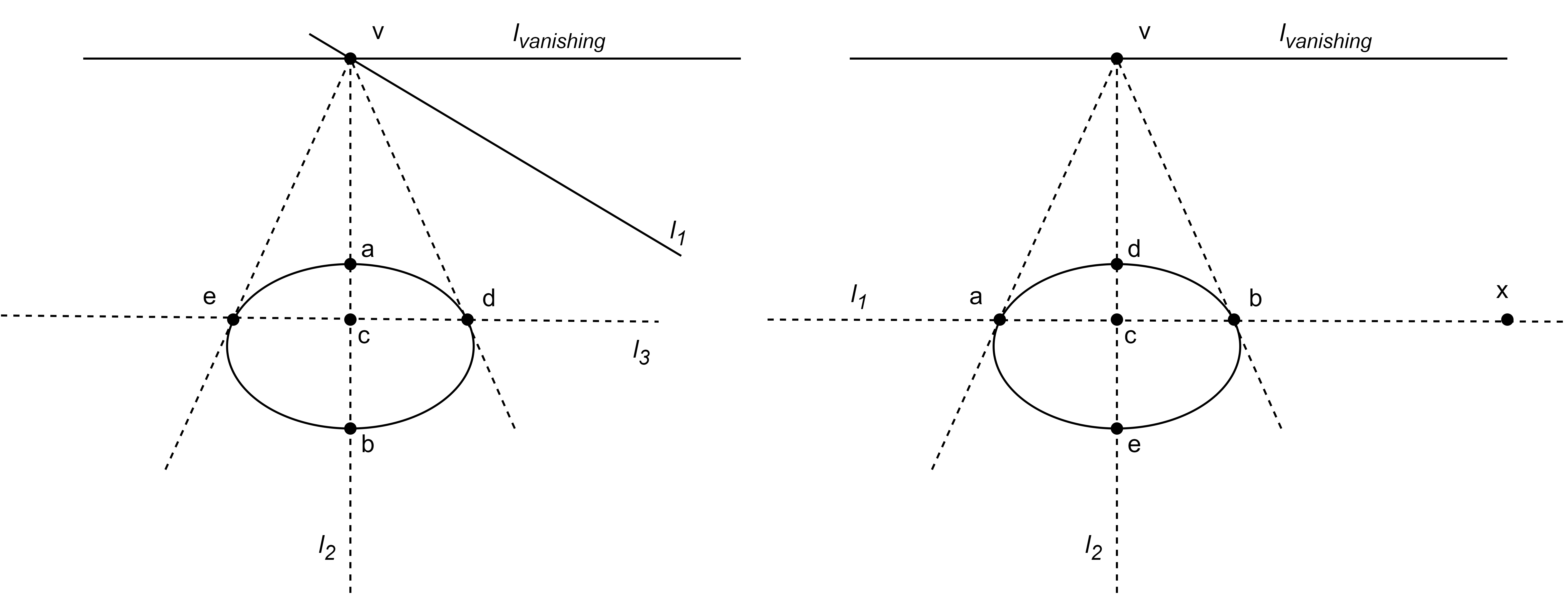}
\end{center}
\caption{Starting either from a line correspondence $\genericLine_1$ (left) or a point correspondence $\vector x$ (right), we derive the dashed lines and the set of points $\vector a$, $\vector b$, $\vector d$, and $\vector e$.  \label{fig:derived_corresp}}
\end{figure*}

Given a point correspondence $\vector x$, we can derive the line that passes through that point and the projected center of the circle: 
\begin{equation}
\label{eq:case2eq1}
\genericLine_1 = \vector x \times  \vector c \dot
\end{equation}
As can be seen in the right part of \cref{fig:derived_corresp}, besides intersecting the ellipse in two points $\vector a$ and $\vector b$, the line $\genericLine _1$ is in a pole-polar relationship with a point on the vanishing line of the plane 
\begin{equation}
\vector v = \matrix E \, \genericLine_1\dot
\end{equation}
As $\genericLine_1$ intersects the ellipse $\matrix E$ in two points, the tangents to these two points $\vector a$ and $\vector b$ intersect at $\vector v$.
The point at infinity $ \vector v$ is the vanishing point of the parallels to the tangents of $\vector a$ and $\vector b$, and can be used to retrieve another parallel that passes through the projected circle center : 
\begin{equation}
\genericLine_2 =  \vector v \times \vector c\comma
\end{equation}
 that intersects the ellipse in two points $\vector d$ and $\vector e$ and, as the tangents of $\vector a$ and $\vector b$ are parallel to $\genericLine_2$, the line $\genericLine_2$ is a perpendicular to $\genericLine_1$. The exact position of the derived points can be easily retrieved from the known position of $\vector x$ and the circle in the sports field template.

While this method works for a generic registration case, its application to soccer is very limited, as the only point correspondences available in central views  all lie on the middle line. In this instance, the first \Cref{eq:case2eq1} deriving a line through the point correspondence and the circle center retrieves the equation of the middle line, and thus the method falls back on the first case of \cref{section:line-correspondence}. We will thus not investigate this method further in our paper.

\subsection{Case 3: Unknown center} \label{sec:hidden-center}
In some cases, the detection of the center of the circle might fail, due to occlusion with players, poor markings' definition in the image, or its absence in the sports field markings. While projective transformations deform a circle into an ellipse, the center of the ellipse is not the projection of the center of the original circle. The estimation of the original center of the circle is not straightforward, even if the ellipse equation is known in the image.

To recover the imaged center, we leverage the fact that the field markings have a consistent thickness of about ten centimeters and can be considered defining two concentric circles at their interface with the ground. 

Once ellipses are detected in the image, we rely on a result of Huang \emph{et al.~}\cite{Huang2015TheCommon} to retrieve the imaged center. Let $\matrix C_1$ and $\matrix C_2$ be the concentric circles, and let $\matrix E_1$ and $\matrix E_2$ be their respective images. These circles share a common pole-polar, as their common center maps the vanishing line of the supporting plane $\pi$. In their work, Huang \etal showed that the image of the circles' center $\vector c=\matrix H \, \vector o$ is an eigenvector of the $\inverse{\matrix E_{2}}\matrix E_{1}$ matrix. 
By estimating the ellipses' coefficient matrices $\matrix E_{1}$
and $\matrix E_{2}$, we first derive $\vector c$ as it is one of the eigenvectors
of $\inverse{\matrix E_{2}}\matrix E_{1}$. Then, $\lvanishing$ is
obtained as the polar of $\vector c$: $\lvanishing=\matrix E_{1}\, \vector c= \lambda \matrix E_{2}\,\vector c$. Subsequently, depending on the support present in the image, we can derive a sufficient number of correspondences by following the methods developed in \cref{section:line-correspondence,section:pt-correspondence}.

\vspace{2mm}
Eventually, for all the aforementioned cases, the ambiguity about left-right and up-down correspondences on the circle are solved based on priors on the camera position. Note that the trapezoid formed by the tangents of the points $\vector a$, $\vector b$, $\vector d$, and $\vector e$, has diagonals that intersect the ellipse in the green points from \cref{fig:top-view-model}. In total, our method gives 8 point correspondences, which can be augmented by 8 lines from their tangents to the ellipse.

\section{Experiments} \label{sec:benchmarks}
In \cref{subsec:ood-experiment}, we will first validate our hypothesis that the previously mentioned virtual augmentation of keypoints method fails to generalize to zoomed-in on the center circle, out-of-domain views. In \cref{subsec:quali,subsec:quanti}, we assess the quality of our method from a circle and a line correspondence. Finally, in \cref{subsec:concentric-experiments}, we assess the robustness of the variant of our method that can deal with missing circle center detection. As our method consists of a correspondence extractor, we will combine it with the calibration algorithm of PnLCalib ~\cite{Gutierrez2024Pnlcalib-arxiv} to obtain the final calibrations.
\subsection{PnLCalib keypoint detector}
\label{subsec:ood-experiment}

A hypothesis on which this work relies is that virtual keypoint neural networks will fail to generalize to out-of-domain (OOD) images. In this experiment, we evaluate the keypoint detector of the best performing camera calibration method on SoccerNet, PnLCalib~\cite{Gutierrez2024Pnlcalib-arxiv}. 
For this experiment, we gathered an OOD test set of 300 central views of empty, smaller stadiums, with lower viewpoints, without ground truth. We make comparisons with the SoccerNet dataset (SN), which consists of the images of the test sets of the \textit{sn-calibration} and \textit{sn-gamestate} datasets where i) the central circle is visible, and ii) the number of visible lines is lower than 3; this test set consists of $8{,}565$ images.
A way to validate the correctness of predicted keypoints without ground truth is to verify the alignment of points that are supposed to belong to the same line. 

\begin{figure}
\begin{centering}
\includegraphics[width=0.95\columnwidth]{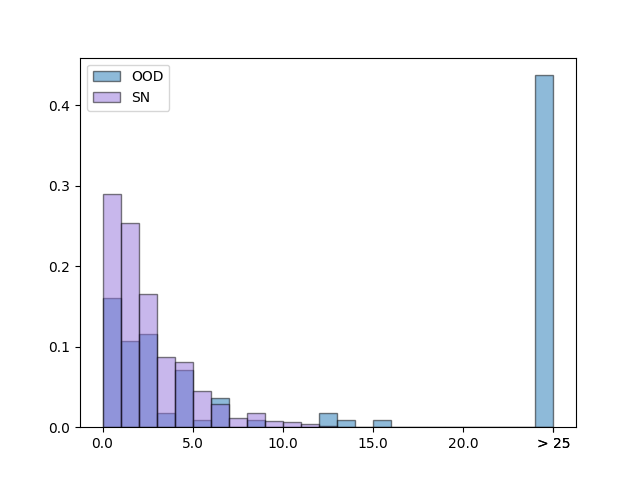}
\par\end{centering}
\caption{Distribution of Euclidean distances, in pixels, between the circle center and each line formed by opposed keypoints pairs. Compared to SoccerNet (SN), the out-of-distribution dataset (OOD) shows a significant number of unbounded errors.\label{fig:distance-line-kp-ood}}
\end{figure}

There are 8 points predicted by PnLCalib neural network on the center circle, of which we discard the points that are the intersection of the center circle and the middle line (red points in \cref{fig:top-view-model}). Then, from each pair of opposed predicted points on the circle, we estimate a line. The center of this circle is supposed to belong to this line, and thus we measure the distance between the center and its projection on the line to assess the soundness of the predicted keypoints. 
The results are depicted in \cref{fig:distance-line-kp-ood}. While the error distribution appears bounded on the SoccerNet dataset, the out-of-distribution images are more subject to high error rates. 
By construction, our geometric method derives pairs of points on the ellipse that are collinear with the circle center.

\subsection{Qualitative results}
\label{subsec:quali}
\begin{figure*}[tbp]
    \centering
    \begin{tabular}{cccc}
       
        \begin{subfigure}[b]{0.22\textwidth}
            \centering
            \includegraphics[width=\textwidth]{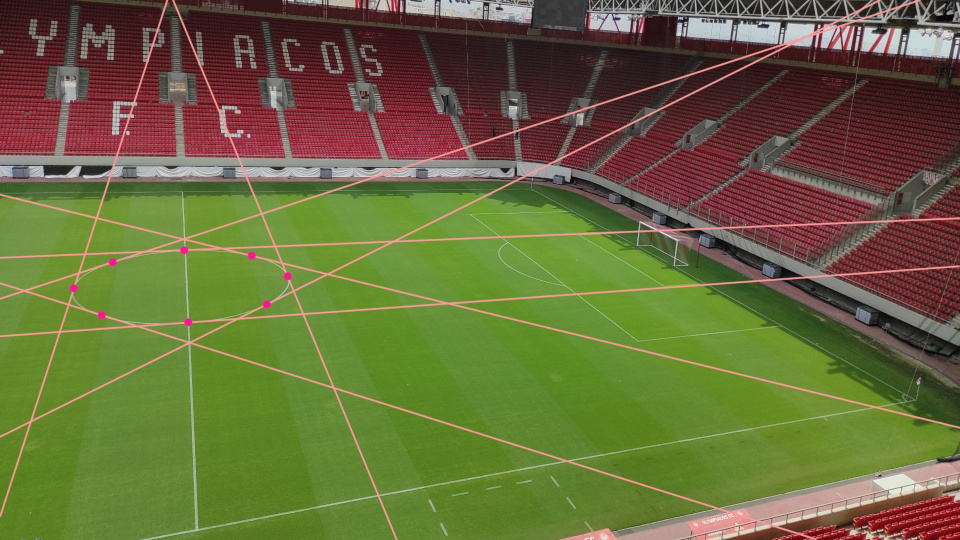}
        \end{subfigure} &
        
         \begin{subfigure}[b]{0.22\textwidth}
            \centering
            \includegraphics[width=\textwidth]{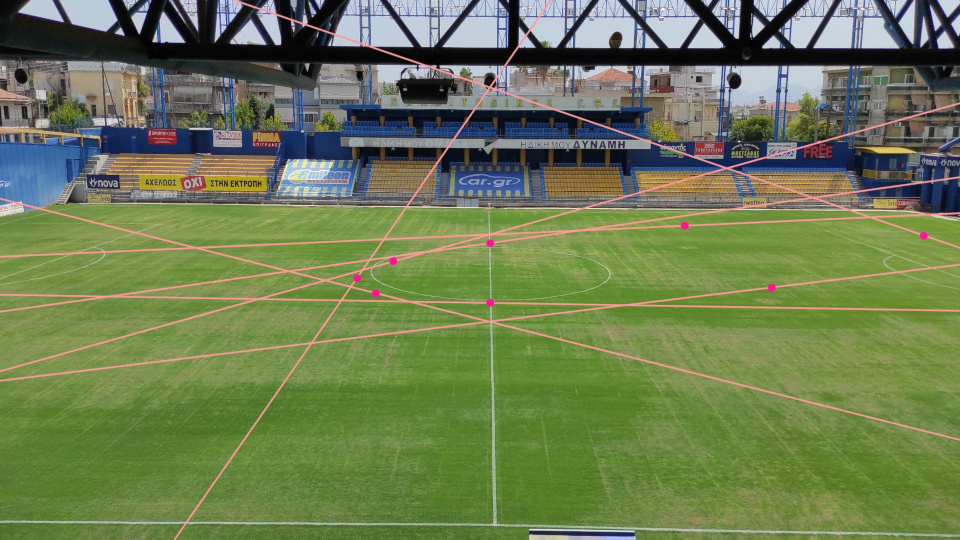}

        \end{subfigure} &
        \begin{subfigure}[b]{0.22\textwidth}
            \centering
            \includegraphics[width=\textwidth]{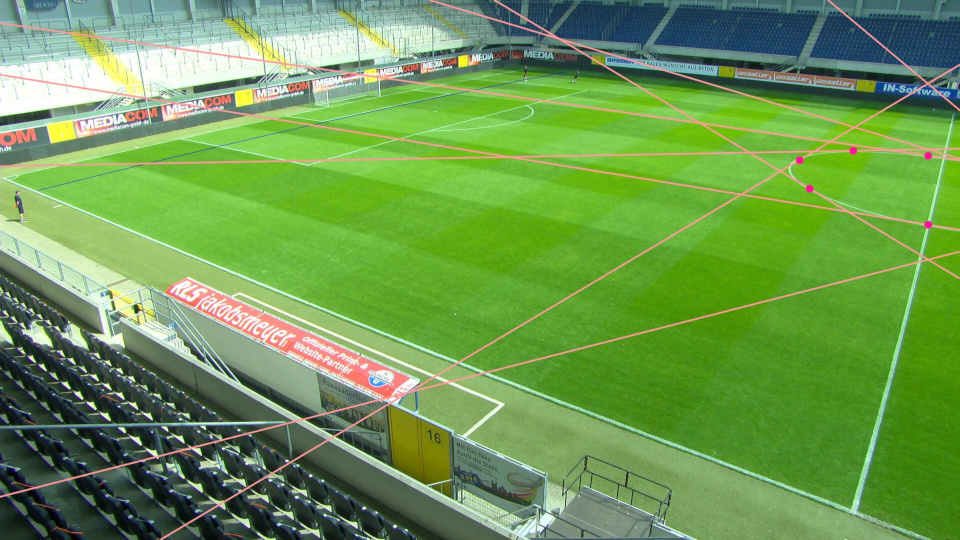}

        \end{subfigure} &
         \begin{subfigure}[b]{0.22\textwidth}
            \centering
            \includegraphics[width=\textwidth]{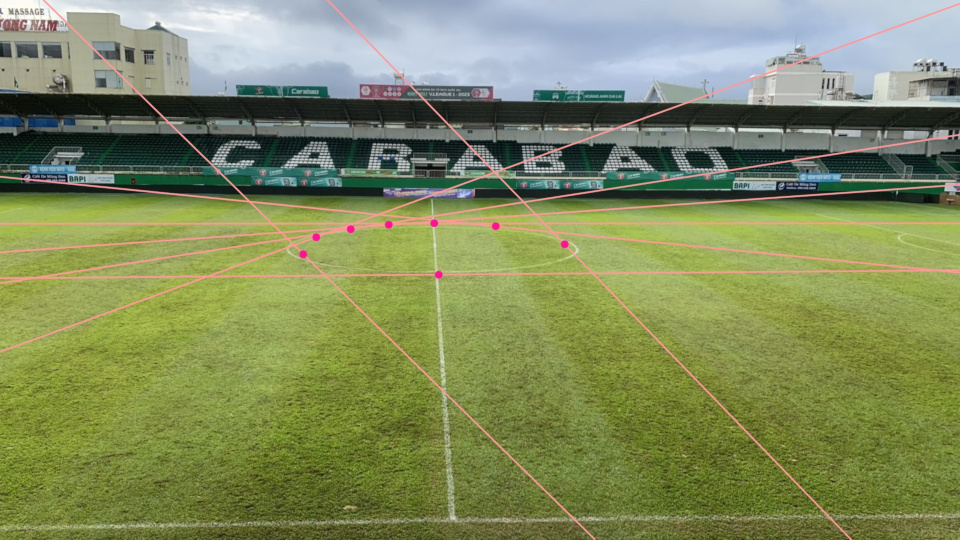}

        \end{subfigure} \\
        \begin{subfigure}[b]{0.22\textwidth}
            \centering
            \includegraphics[width=\textwidth]{keypoints__01130.jpg}

        \end{subfigure} &

        \begin{subfigure}[b]{0.22\textwidth}
            \centering
            \includegraphics[width=\textwidth]{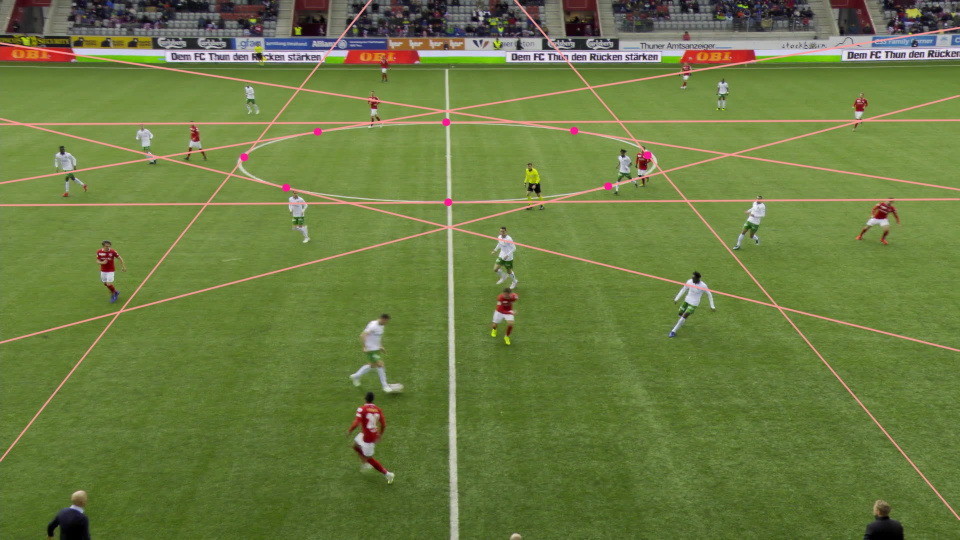}
        \end{subfigure} &
         \begin{subfigure}[b]{0.22\textwidth}
            \centering
            \includegraphics[width=\textwidth]{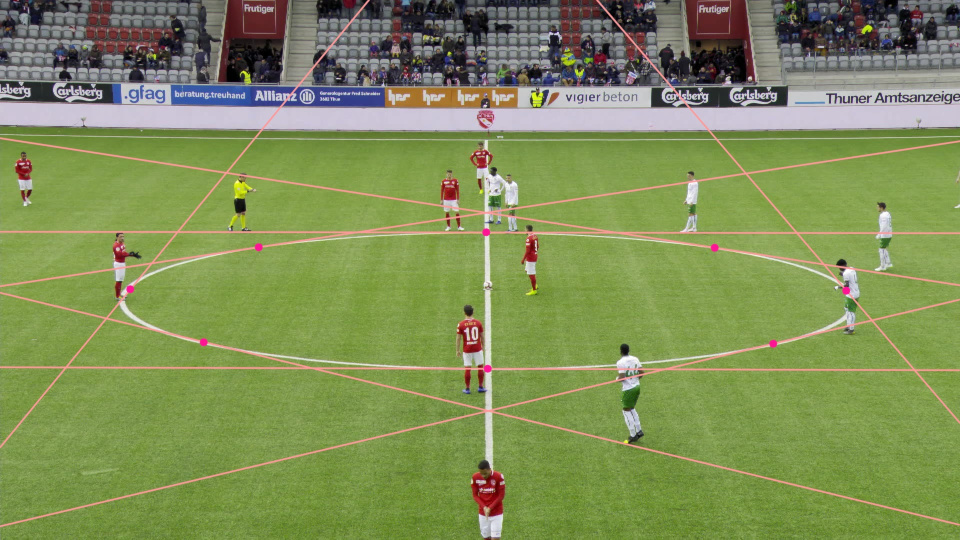}

        \end{subfigure} &
        \begin{subfigure}[b]{0.22\textwidth}
            \centering
            \includegraphics[width=\textwidth]{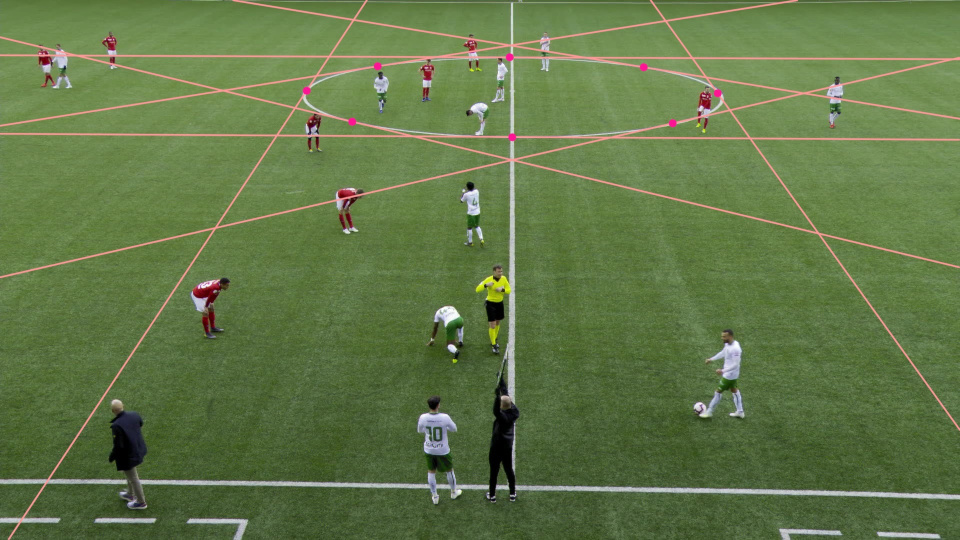}

        \end{subfigure} \\

         \begin{subfigure}[b]{0.22\textwidth}
            \centering
            \includegraphics[width=\textwidth]{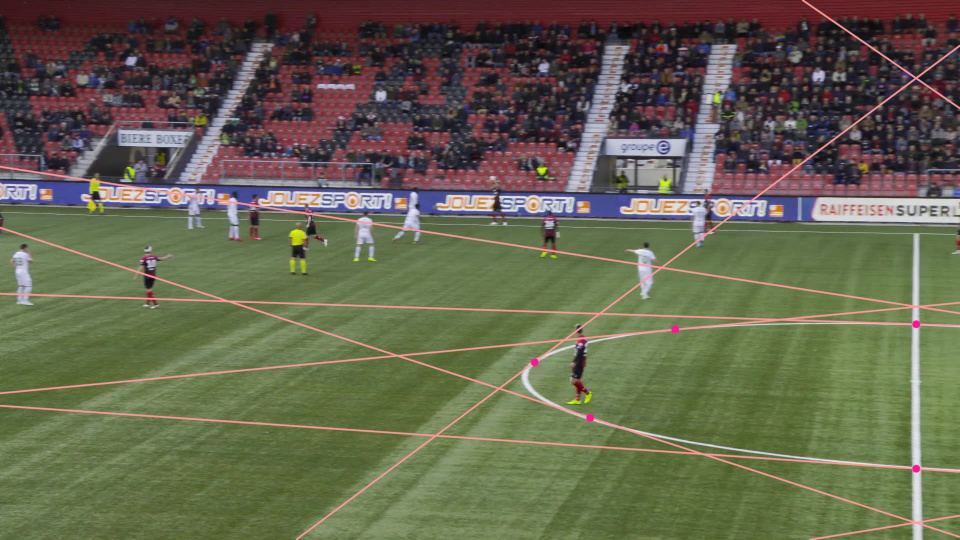}

        \end{subfigure} &
        \begin{subfigure}[b]{0.22\textwidth}
            \centering
            \includegraphics[width=\textwidth]{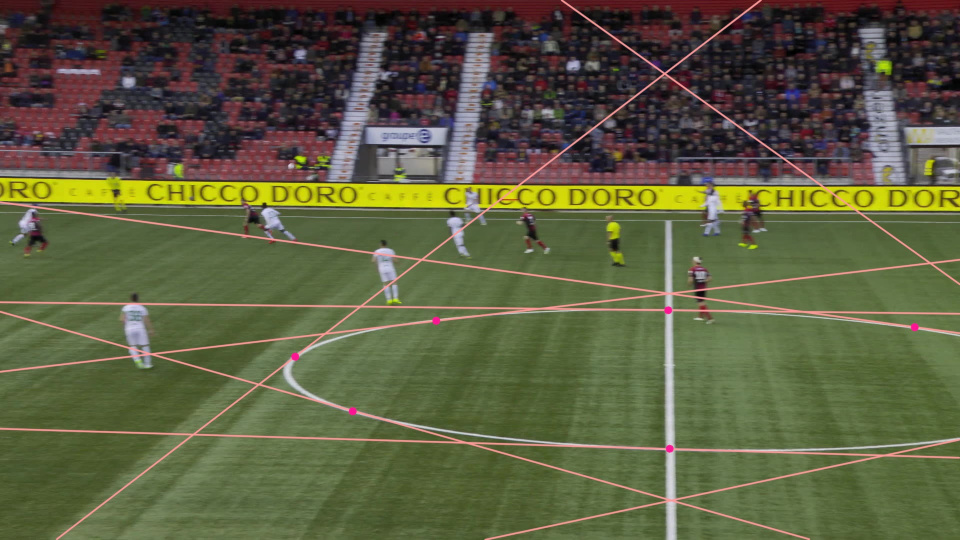}

        \end{subfigure} &

        \begin{subfigure}[b]{0.22\textwidth}
            \centering
            \includegraphics[width=\textwidth]{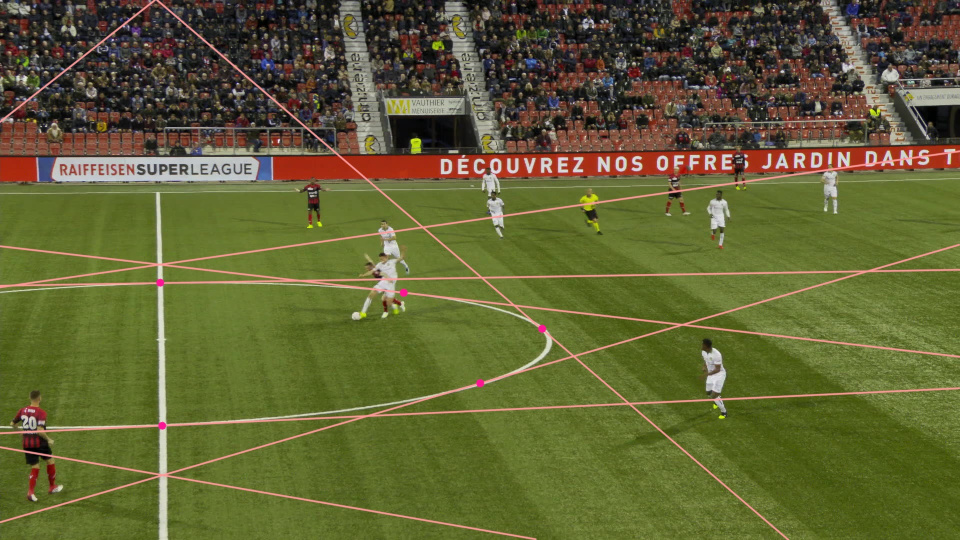}
        \end{subfigure} &
         \begin{subfigure}[b]{0.22\textwidth}
            \centering
            \includegraphics[width=\textwidth]{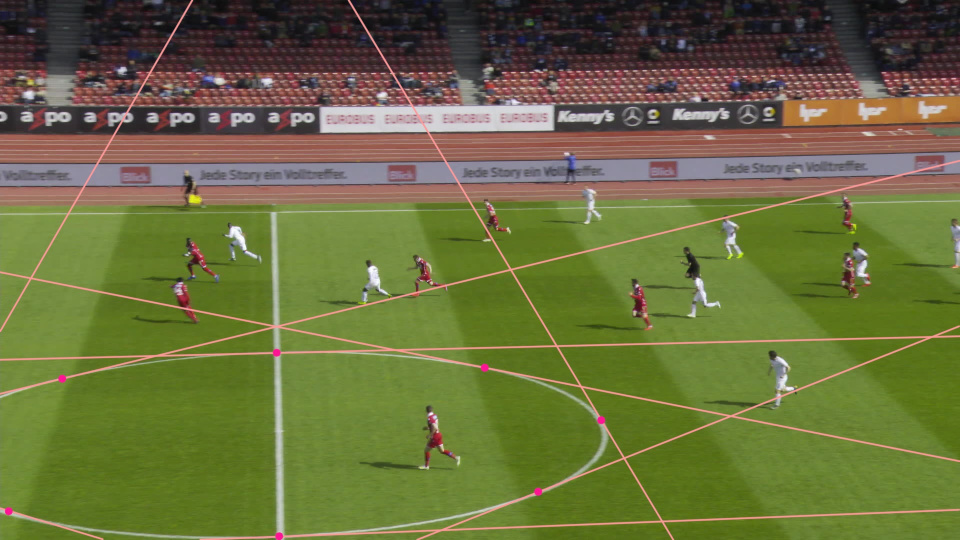}

        \end{subfigure} 
        \\
    
    \end{tabular}
    \caption{Various results of our method extracting lines and points from the center circle and the middle line correspondences. The images come from either our OOD dataset, or SoccerNet. The few failures are due to poor ellipse fitting or to bad center detection. }
    \label{fig:retrieved-points}
\end{figure*}

\begin{figure*}[tbp]
    \centering
    \begin{tabular}{cccc}
       
        \begin{subfigure}[b]{0.22\textwidth}
            \centering
            \includegraphics[width=\textwidth]{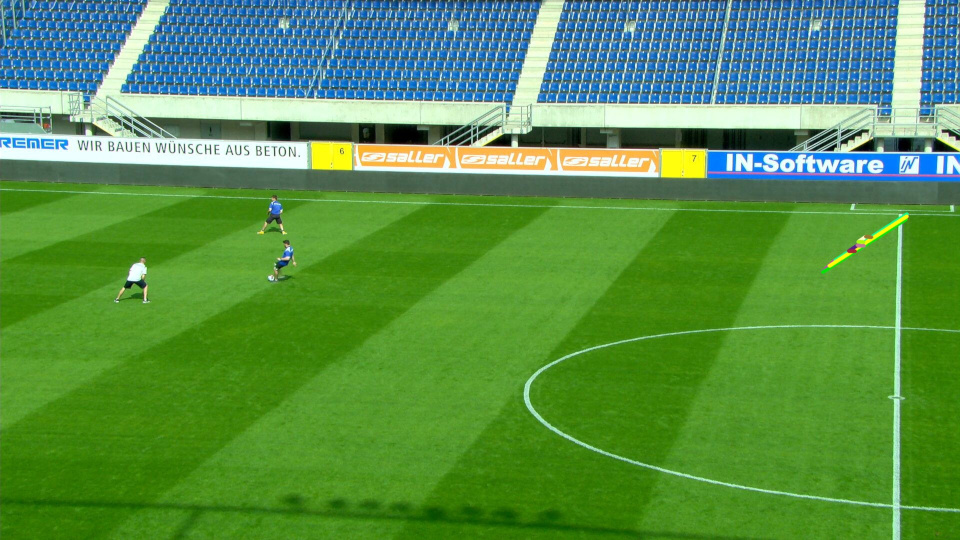}
        \end{subfigure} &
         \begin{subfigure}[b]{0.22\textwidth}
            \centering
            \includegraphics[width=\textwidth]{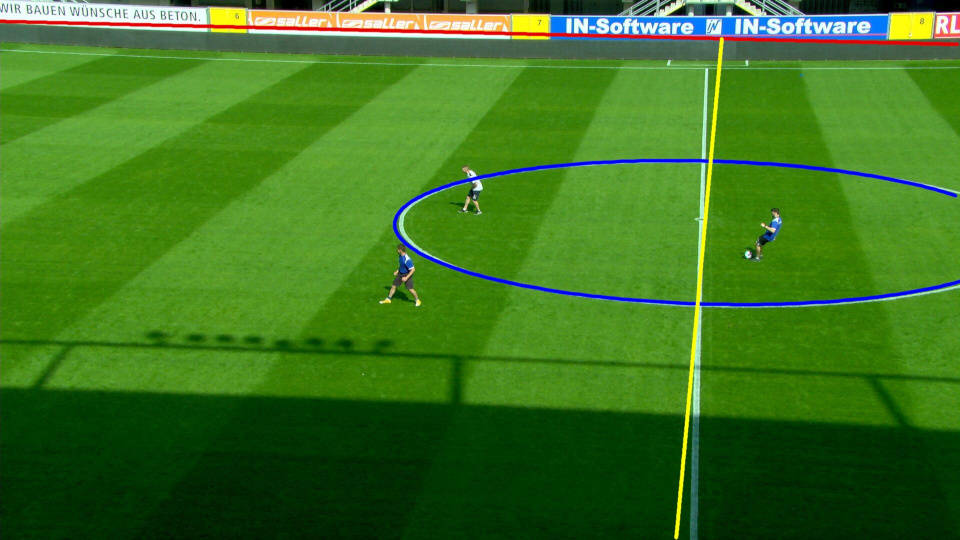}

        \end{subfigure} &
        \begin{subfigure}[b]{0.22\textwidth}
            \centering
            \includegraphics[width=\textwidth]{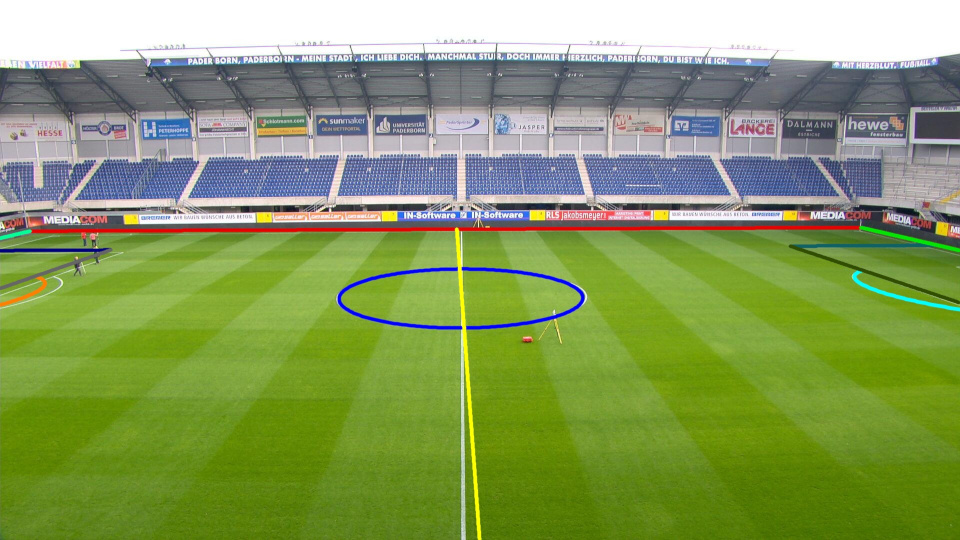}

        \end{subfigure} &
         \begin{subfigure}[b]{0.22\textwidth}
            \centering
            \includegraphics[width=\textwidth]{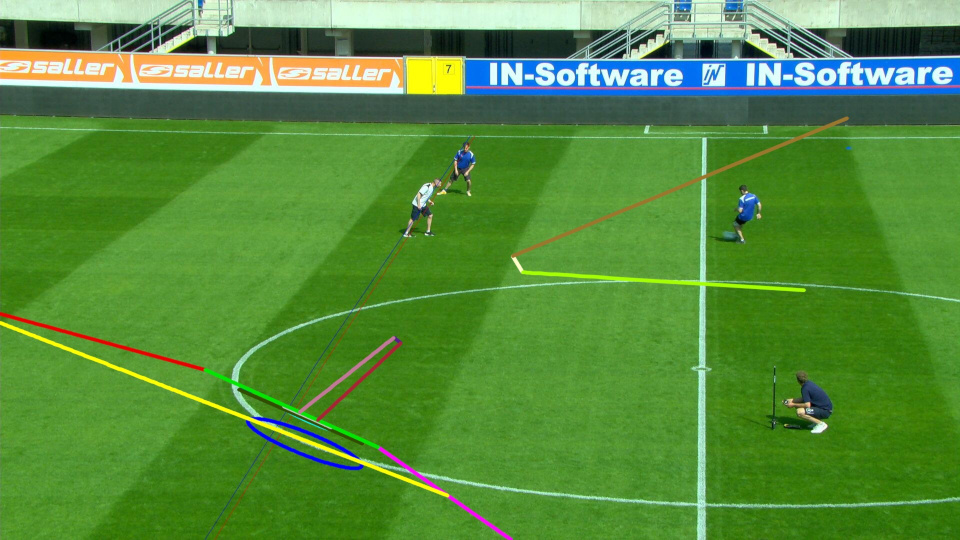}
        \end{subfigure} \\
        
         \begin{subfigure}[b]{0.22\textwidth}
            \centering
            \includegraphics[width=\textwidth]{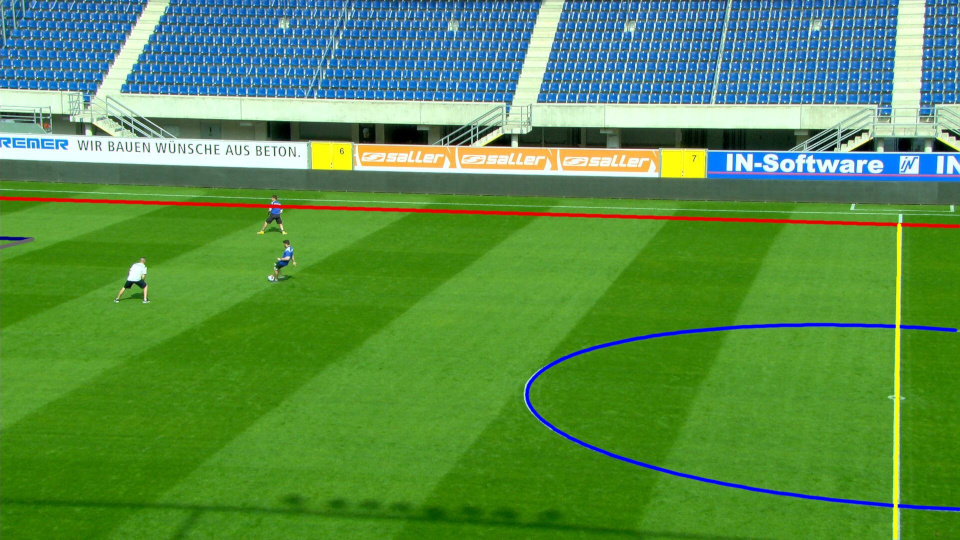}

        \end{subfigure} &
        \begin{subfigure}[b]{0.22\textwidth}
            \centering
            \includegraphics[width=\textwidth]{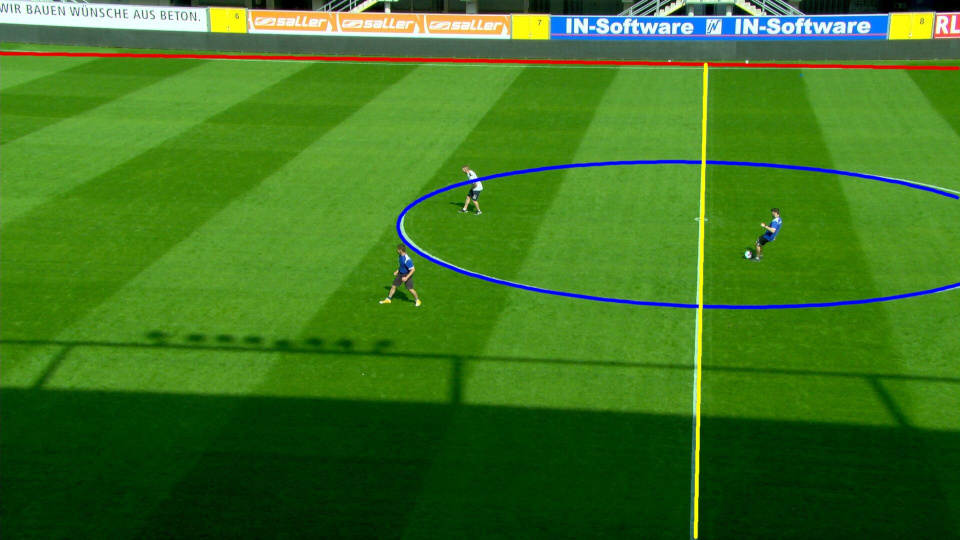}
                    \end{subfigure} 
            &
            
         \begin{subfigure}[b]{0.22\textwidth}
            \centering
            \includegraphics[width=\textwidth]{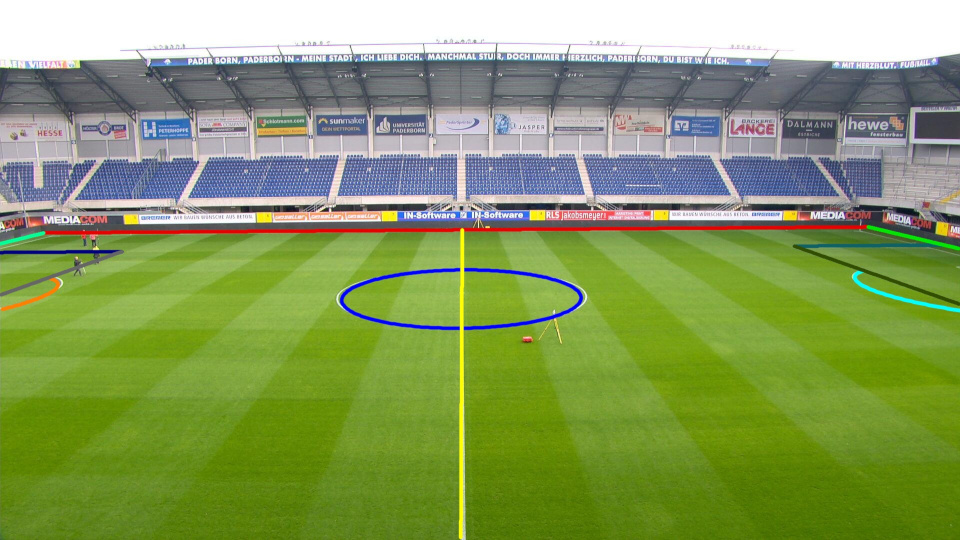}

        \end{subfigure} &
        \begin{subfigure}[b]{0.22\textwidth}
            \centering
            \includegraphics[width=\textwidth]{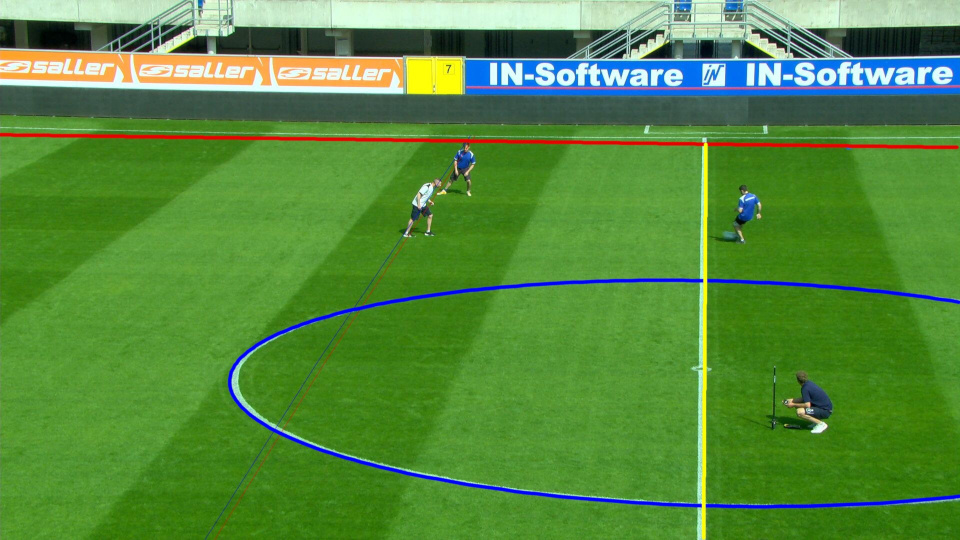}

        \end{subfigure}

    \end{tabular}
    \caption{Top-row: vanilla PnLCalib results on OOD frames. Bottom-row: PnLCalib calibration when our keypoints are included.}
    \label{fig:PnL-corrected}
\end{figure*}

Starting from the semantic segmentation detector of TVCalib~\cite{Theiner2023TVCalib}, we estimate the ellipse parameters. The center and the intersection of the center circle with the middle line are estimated thanks to the keypoint detector of PnLCalib~\cite{Gutierrez2024Pnlcalib-arxiv}. From these detections, we derive point and line correspondences using the method described in \cref{section:line-correspondence}. A series of results are displayed in \cref{fig:retrieved-points}, while the effect of including these points in a full camera calibration pipeline is depicted in \cref{fig:PnL-corrected}. The results shown in \cref{fig:retrieved-points,fig:PnL-corrected} indicate that, if the initial detector that provides the circle correspondence and the distinguishable keypoints are performing well, our geometric approach gives promising results, enabling to calibrate cameras in challenging cases, or improving the estimated perspectivity as shown in \cref{fig:PnL-corrected}. 

\subsection{Quantitative Results}
\label{subsec:quanti}

The difficulty to showing the benefits of our methods using a quantitative analysis is that, as previously mentioned, it tackles mainly out-of-domain images that are by definition not in the annotated datasets. For the sake of completeness, however, we conducted a quantitative analysis on the SoccerNet dataset.  For this setup, given the validation of the predicted keypoint correctness on the SoccerNet dataset of \cref{subsec:ood-experiment}, we do not expect our geometric approach to change the results of PnLCalib when we replace their predicted keypoints by our geometric keypoints. The results are reported in \cref{tab:metrics}, and as expected, no significant change is to be observed with our keypoints.

\begin{table}
    \centering
 
     \resizebox{1.0\columnwidth}{!}{
     \begin{tabular}{c|ccc} 
         & $\calibrationMetric_{5}~(\uparrow)$ [\%]  &  $\meanReprojectionError~(\downarrow)$ [px] & CR$~(\uparrow)$ [\%] \\ \hline
         TVCalib~\cite{Theiner2023TVCalib} & 10.2 & 28.8 & \textbf{100}\\
         PnLCalib~\cite{Gutierrez2024Pnlcalib-arxiv} & \textbf{22.4} &  \textbf{12.8}  & 79.5\\
         PnLCalib* & \textbf{22.4} & 12.9  & 79.5\\
    \end{tabular}
    }
      \caption{Jaccard Index for camera calibration at 5 pixels ($\calibrationMetric_{5}$) ~\cite{Magera2024AUniversal} in full-HD frames, Mean Reprojection Error (MRE) in pixels and Completeness Rate (CR) of previous works and PnLCalib* tuned with our geometrically derived keypoints instead of their learned version, evaluated on the central views of SoccerNet test sets.}
    \label{tab:metrics}
\end{table}

\subsection{Unknown center}
\label{subsec:concentric-experiments}
We will now assess our method that allows calibration even when the circle center is not detected in the image. In this experimental setup, we assess our method independently of sports field markings detectors.

\mysection{Experimental setup.} We randomly generate plausible views of soccer main cameras. To do so, we sample common camera positions, orientations and focal length values, and keep the images displaying the center circle and the halfway line only. The pinhole parameters of these cameras are then converted to homographies mapping the synthetic image to the top-view soccer field template with the playfield located in the plane $Z=0$. These ground-truth homographies $\matrix H$ are compared to the estimated homographies $ \hat{\matrix H}$ through the reprojection error of the visible 3D points belonging to the soccer field in the synthetic image.
Specifically, given a set of 3D visible points $\{\textbf{x}_{1},...,\, \textbf{x}_{N}\}$  in a synthetic frame, we define the mean reprojection error in the image as: 
$\meanReprojectionError = \dfrac{1}{N} \sum_{i}^{N} \left\Vert \matrix H \, \textbf{x}_i- \hat{\matrix H} \textbf{x}_i \right\Vert _{2} \dot
$

\begin{figure*}[ptb]
    \centering
    \begin{tabular}{cc}
      
        \begin{subfigure}[b]{0.47\textwidth}
            \centering
            \includegraphics[width=\textwidth]{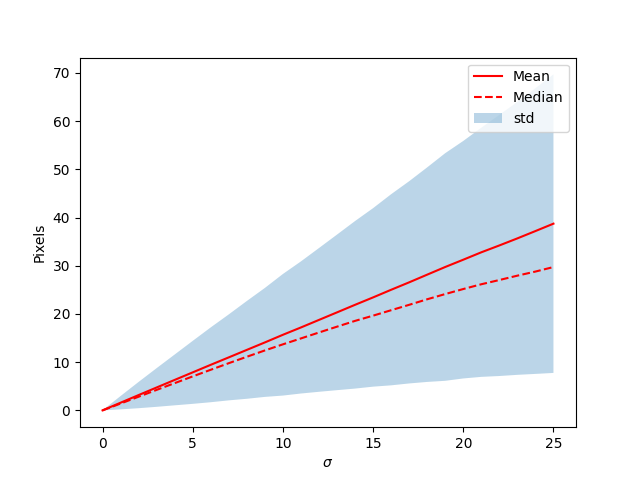}
            \caption{}
            \label{fig:syn-noise-ellipses}
        \end{subfigure} &
        \begin{subfigure}[b]{0.47\textwidth}
            \centering
            \includegraphics[width=\textwidth]{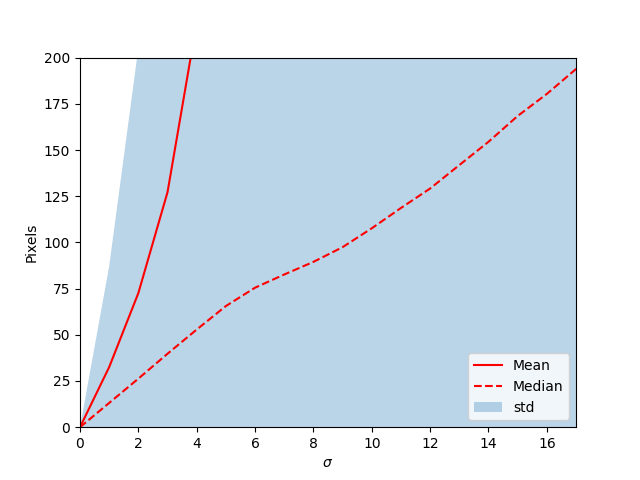}
            \caption{}
            \label{fig:syn-noise-center}

        \end{subfigure} 
    \end{tabular}
    \caption{Reprojection errors when points are modified with a Gaussian noise with zero mean and standard deviation $\sigma$. (a) corresponds to an increasing noise applied to ellipse points, and (b) to an increasing noise applied to the circle center estimate.\label{fig:Synthetic-noise}}
\end{figure*}

\mysection{Results on synthetic data.}
\label{subseubsec:unknown-center-synthetic-exp}
In our simulations, assuming a field marking thickness of 8 centimeters, we sample 8 points along the inner and outer ellipses corresponding to the central concentric circles. Those points are projected in the full-HD (1080p) image, and a zero-centered Gaussian noise with increasing standard deviation $\sigma$ is applied independently to each point. The ellipses are fitted with Fitzgibbon's method~\cite{Fitzgibbon1995ABuyer}, then our method is applied to derive the final homography. In our synthetic data experiments, we vary the standard deviation $\sigma$ between 0 and 25 pixels for 100 randomly generated cameras. As it can be observed in~\cref{fig:syn-noise-ellipses}, we find that the level of error seems to linearly follow the noise applied to the ellipse points. In a second experiment, we measure the sensitivity of our method to the location of the circle center. To do so, we apply the same noise to the 2D position of the center. \Cref{fig:syn-noise-center} shows that the reprojection error rises rapidly with this type of noise, hence highlighting the need for a precise center estimation. 

\begin{figure*}[tbp]
    \centering
    \begin{tabular}{ccc}
       
        \begin{subfigure}[b]{0.31\textwidth}
            \centering
            \includegraphics[width=\textwidth]{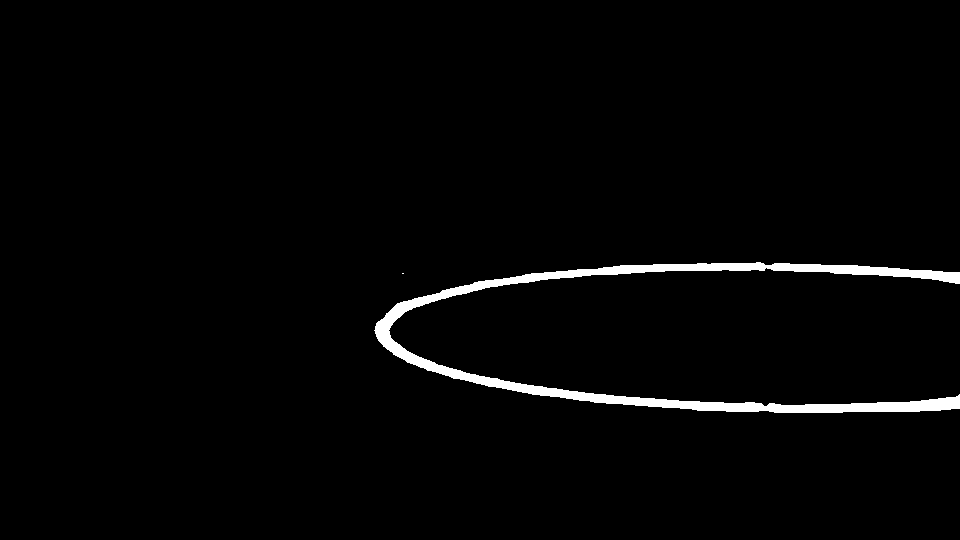}
            \caption{}
        \end{subfigure} &
         \begin{subfigure}[b]{0.31\textwidth}
            \centering
            \includegraphics[width=\textwidth]{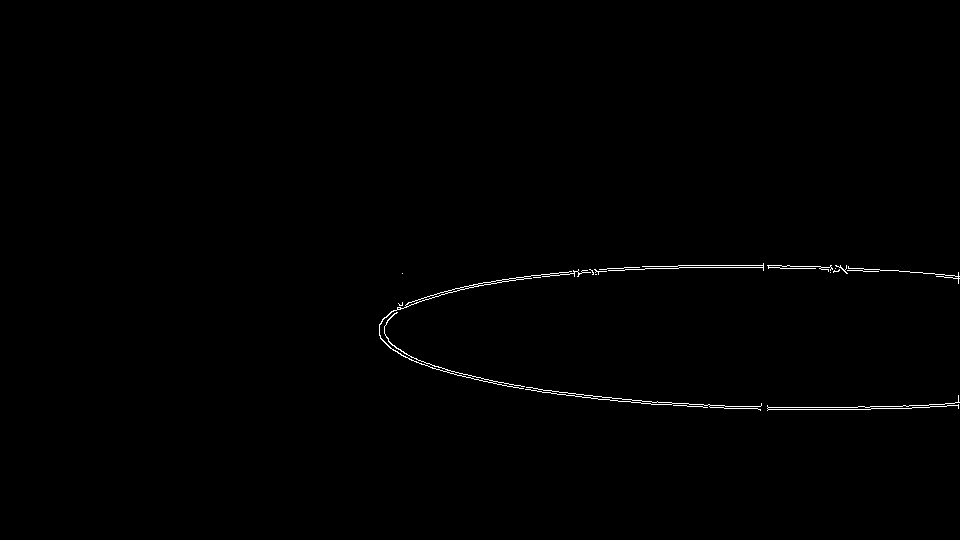}
                        \caption{}

        \end{subfigure} &
        \begin{subfigure}[b]{0.31\textwidth}
            \centering
            \includegraphics[width=\textwidth]{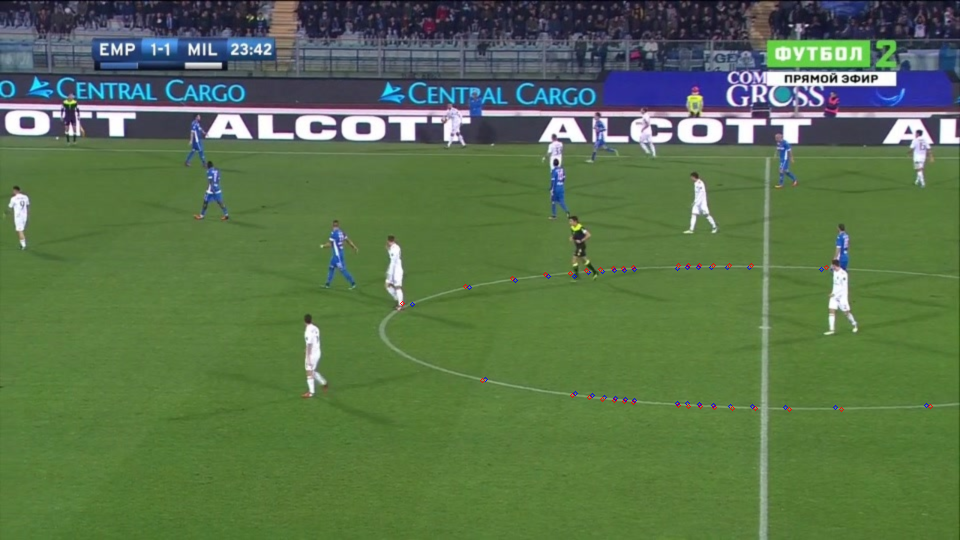}
                        \caption{}

        \end{subfigure}

    \end{tabular}
    \caption{Concentric circles detection. (a) displays the semantic segmentation of the ``Center circle'' class of the soccer field, (b) shows the result of Canny's detector in the segmented area, and (c) shows inliers points for inner (blue) and outer (red) ellipse points.}
    \label{fig:concentric circles detection}
\end{figure*}

\mysection{Results on real data.} The existing sports circular markings detectors do not detect both inner and outer edges of the markings. Usually, they perform semantic segmentation, and neglect the field markings thickness. To detect both edges, we first use the semantic segmentation of the circle by using Theiner's line segmentation neural network~\cite{Theiner2023TVCalib}, which determines a rough local area to search for inner and outer ellipses. To estimate those, we apply Canny's edge detector in the segmented area. We roughly estimate the center of the detected area, and starting from this center, we throw rays that should intersect the detected edges of both ellipses only a few pixels apart. The first detected edges along the rays belong to the candidates for the inner circle, and the second ones belong to the candidates for the outer circle. We use RANSAC to derive the final ellipses' equations in the image. \Cref{fig:concentric circles detection} shows the results of each intermediate step. 
\begin{figure}
    \centering
    \begin{tabular}{cc}
       
        \begin{subfigure}[b]{0.22\textwidth}
            \centering
            \includegraphics[width=\textwidth]{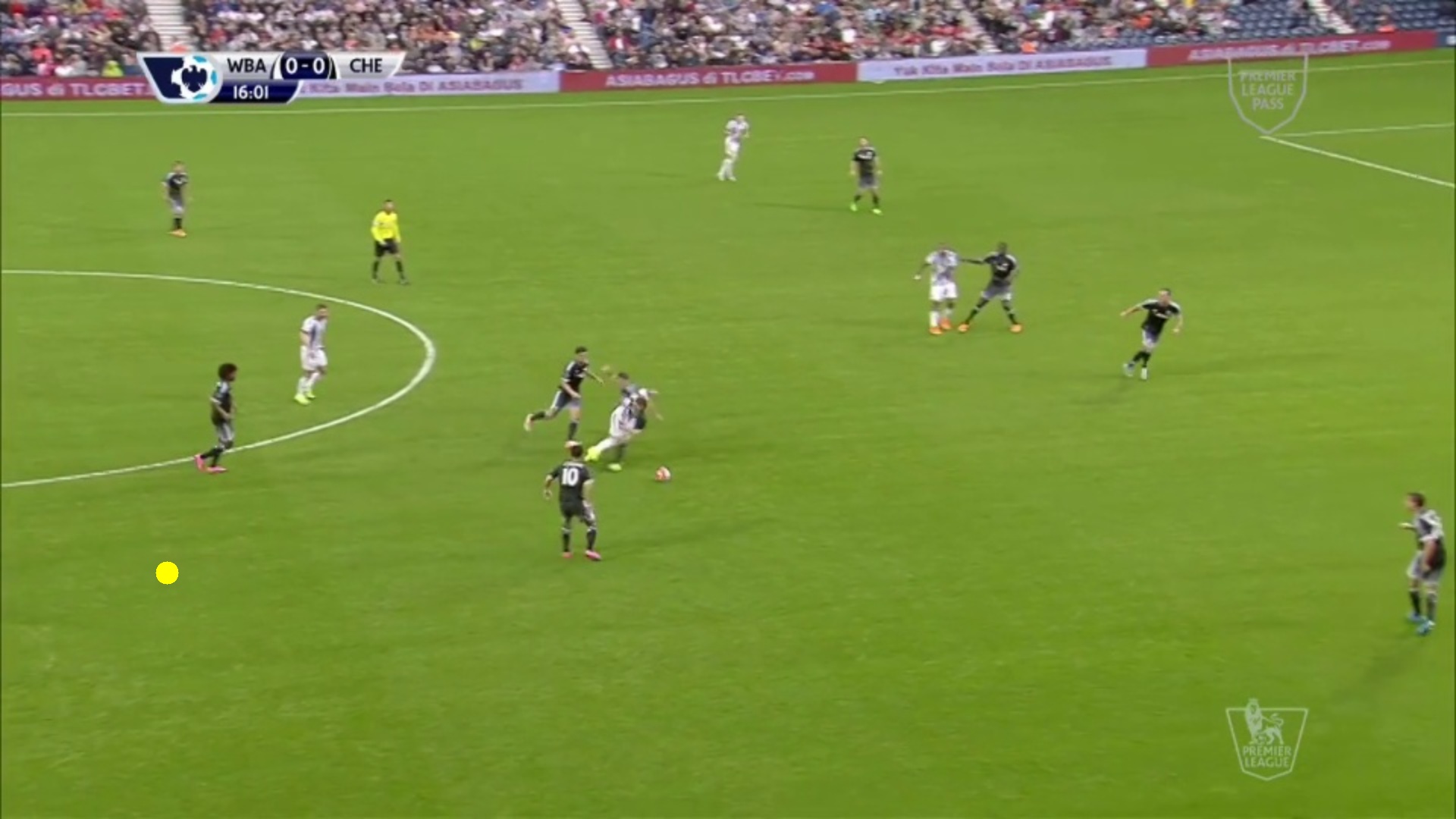}
        \end{subfigure} &
         \begin{subfigure}[b]{0.22\textwidth}
            \centering
            \includegraphics[width=\textwidth]{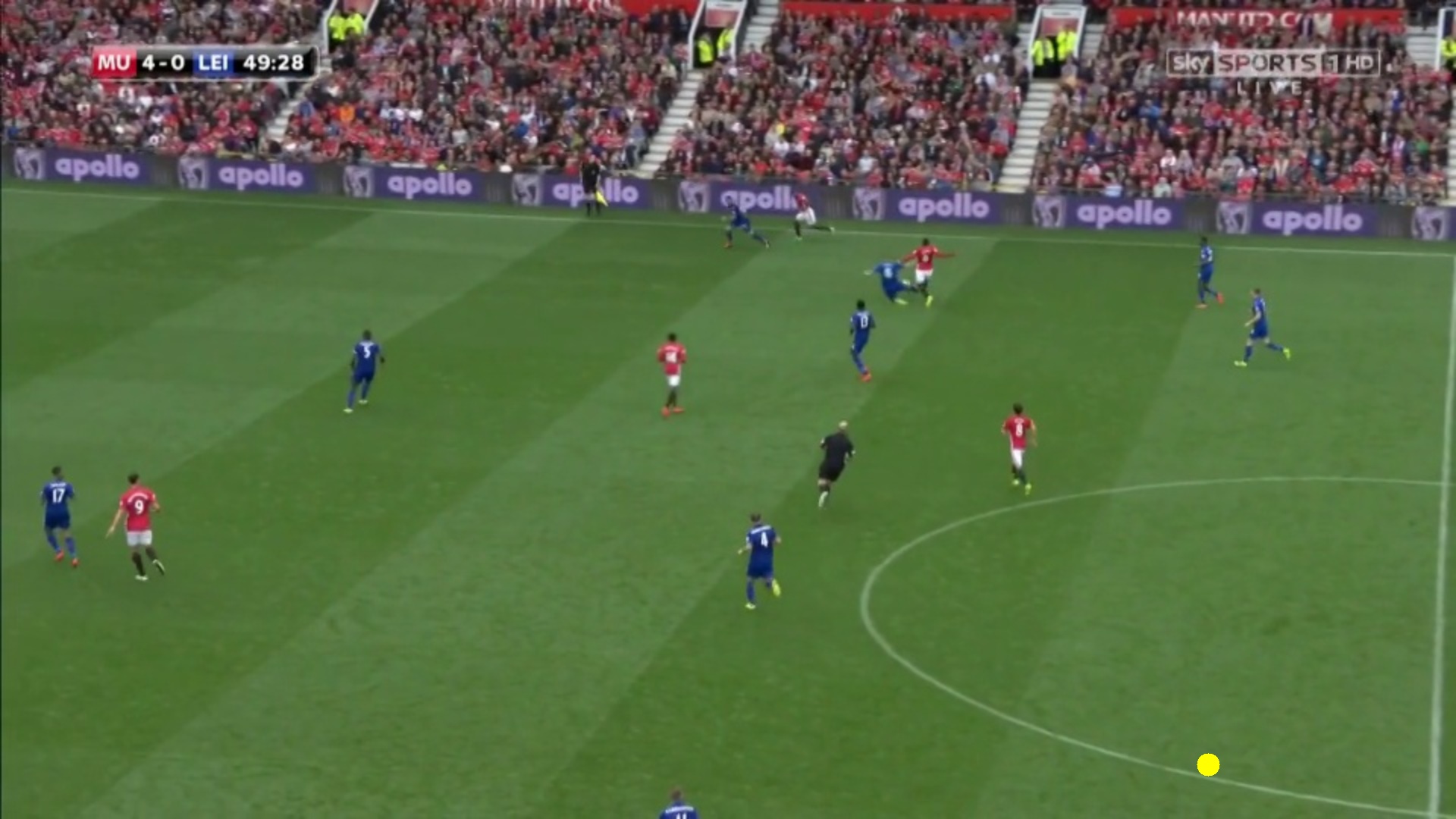}
        \end{subfigure} 
         \\
         \begin{subfigure}[b]{0.22\textwidth}
            \centering
            \includegraphics[width=\textwidth]{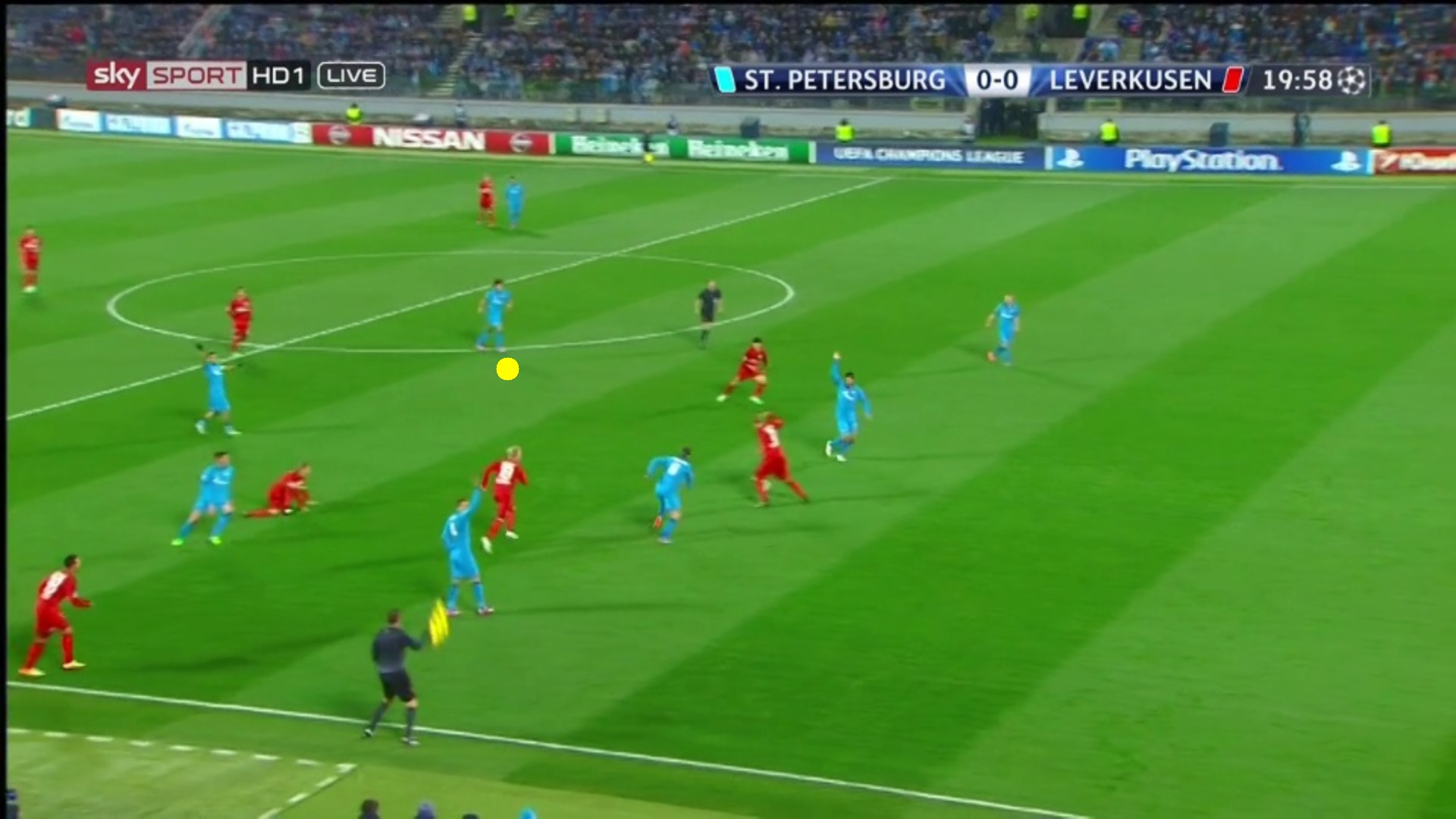}
        \end{subfigure} 
         &
         \begin{subfigure}[b]{0.22\textwidth}
            \centering
            \includegraphics[width=\textwidth]{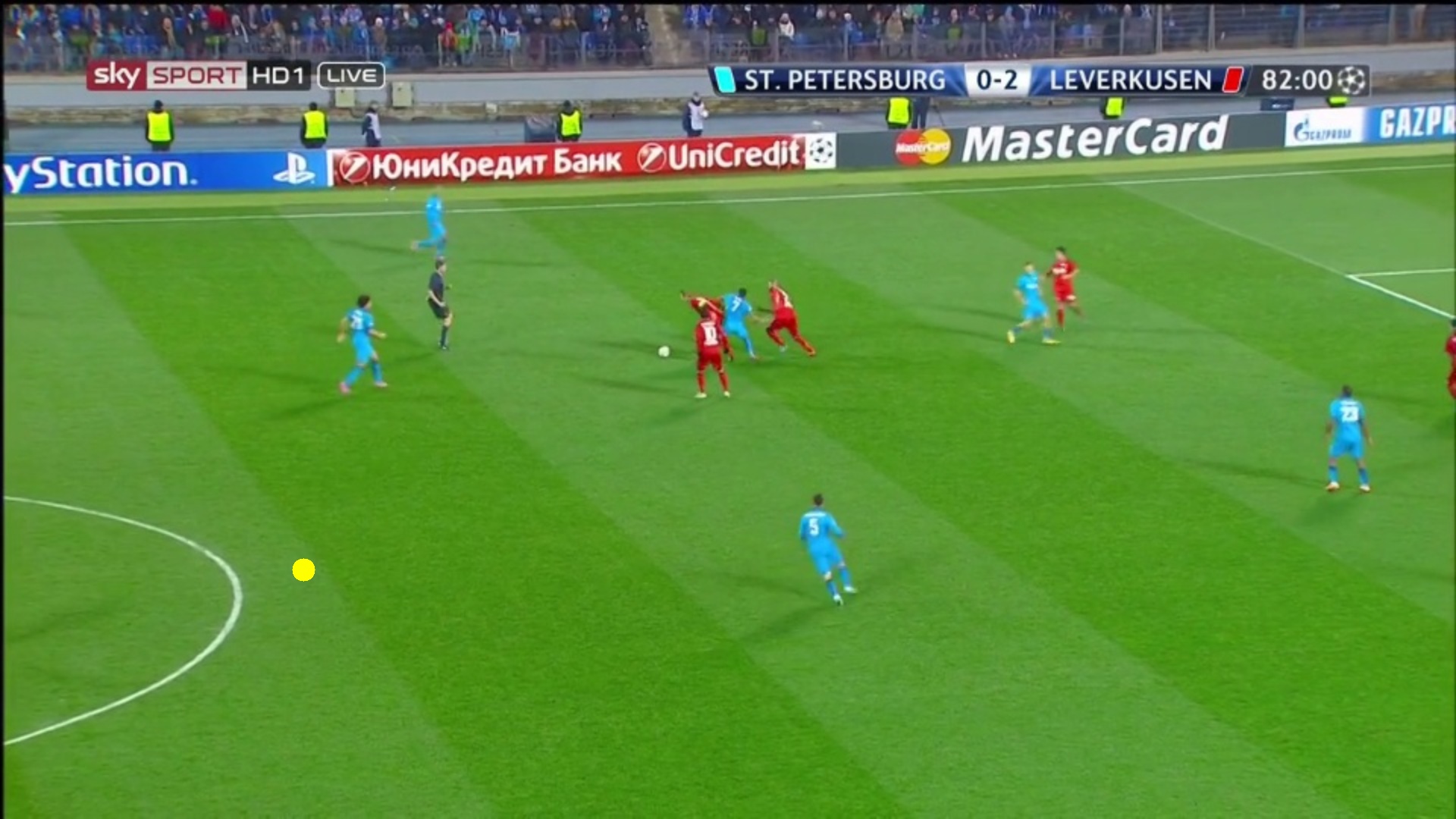}
        \end{subfigure} 
    
    \end{tabular}
    \caption{Circle centers (yellow) detected based on the concentric circles' detection. The results are poor, as synthetic experiments suggest. }
    \label{fig:detected-center-concentric}
\end{figure}
Finally, examples of the center derived from the two ellipses are displayed in \cref{fig:detected-center-concentric}. The retrieved centers are too noisy to be used in real-world scenarios. The views, where only a part of the ellipse is visible, excluding the middle line for instance, remain very challenging views to calibrate or register. This is also true for the views where the center keypoint detector fails, for example when the center is occluded or almost invisible due to poor image quality.

\section{Discussion}
Our contribution to the quality of the annotations of keypoints on circles is obvious, as our method enables to annotate views for which no homography could be derived from points or lines alone. Furthermore, as shown in \cref{fig:bad-tangents}, it considerably improves the  geometric consistency compared to great axis annotation. 
The benefits of our method for calibrating cameras are more difficult to measure, as it tackles mainly out-of-domain images that are by definition not in the annotated datasets. Still, our qualitative experiments indicate that our geometric method can retrieve points and lines correspondences that enable camera calibration in difficult scenarios by relying on the generalization capacities of detectors for actual recognizable patterns such as circular field markings or line intersections.
The main limitations of our method come from the dependence on such robust detectors, which may overfit the few open-source datasets, and from the noise that may arise in ellipse estimation, mainly in the concentric setup where the line thickness is probably insufficient to guarantee a reliable circle center extraction. The cases depicting only a small part of the ellipse require further investigation to enable their calibration.

\section{Conclusion}\label{sec:conclusion}

In this paper, we have shown how to better annotate central views in sports by using a geometric approach. Furthermore, while we show that learned keypoint detectors such as the one of PnLCalib~\cite{Gutierrez2024Pnlcalib-arxiv} perform well to detect keypoints on circular patterns in in-domain scenarios, our geometric approach is effective when the test images are out-of-domain. However, some challenging views remain to be solved for single-frame camera calibration when the central circle center is out of the image or cannot be detected, which we will explore in future work.  

\mysection{Acknowledgments.}
We are grateful to Quentin Massoz and Olaf Wysocki for their valuable inputs.
This work was supported by the Service Public de Wallonie (SPW) Recherche, Belgium, under Grant $\text{N}^{\text{o}}$8573.

\cleardoublepage

\end{document}